%% file: main.tex
\documentclass[runningheads,fleqn]{cl2emult}
\usepackage{makeidx}  
\usepackage{graphicx} 
\usepackage{subeqnar} 
\usepackage{multicol} 
\usepackage{cropmark} 
\usepackage{engi}     
\makeindex            

\usepackage{url,cite,amsmath,epsfig,epsf,latexsym,bbm}

\newcommand{\M}[1]{\mathbf{#1}}
\newcommand{\V}[1]{\mathbf{#1}}
\newcommand{\norm}[1]{\left | \left | #1 \right | \right |}
\newcommand{\RR}{\mathbbm{R}}
\usepackage{picinpar}

\begin{document}

\title*{Online Searching with an Autonomous Robot}
\toctitle{Online Searching with an Autonomous Robot}
\titlerunning{Online Searching with an Autonomous Robot}

\author{S\'andor P. Fekete\inst{1}
\and Rolf Klein\inst{2}
\and Andreas N\"uchter\inst{3}}

\authorrunning{S\'andor P. Fekete et al.}

\institute{
Department of Mathematical Optimization,
Braunschweig University of Technology, D--38106 Braunschweig, Germany.
Email:~\texttt{s.fekete@tu-bs.de}
\and
Institute of Computer Science,
University of Bonn, D--53117 Bonn, Germany.
Email:~\texttt{rolf.klein@uni-bonn.de}
\and
Fraunhofer Institute for Autonomous Intelligent Systems, Schloss Birlinghoven, D--53754 Sankt Augustin, Germany.
Email:~\texttt{andreas.nuechter@ais.fraunhofer.de}
}
\date{}

\maketitle

\input{s1} 

\input{s2} 

\input{s3} 

\input{s4} 

\input{s5} 

\bibliographystyle{plain}
{\footnotesize

\bibliography{refs}
}

\end{document}

%% file: s1.tex

\begin{abstract}
We discuss online strategies for visibility-based
searching for an object hidden behind a corner,
using Kurt3D, a real autonomous mobile robot. This task
is closely related to a number of well-studied problems.
Our robot uses
a three-dimensional laser scanner in a stop, scan, plan, go fashion 
for building a virtual three-dimensional environment.
Besides planning trajectories and avoiding obstacles, Kurt3D
is capable of identifying objects like a chair. 
We derive a practically useful and asymptotically
optimal strategy that guarantees a competitive ratio of 2,
which differs remarkably from the well-studied scenario
without the need of stopping for surveying the environment.
Our strategy is used by Kurt3D, documented in a separate video.
\end{abstract}

{\bf Keywords:} Searching, visibility problems, watchman problems,
online searching, competitive strategies, autonomous mobile robots,
three-di\-men\-si\-onal laser scanning, Kurt3D.

\section{Introduction}
{\bf Visibility Problems.}
Visibility-based problems of surveying, guarding, or searching
have a long-standing tradition in the area of computational optimization;
they may very well be considered a field of their own. 
Using stationary positions for guarding a region
is the well-known {\em art gallery problem}~\cite{o-agta-87}.
The {\em watchman problem}~\cite{thi-iacsw-93,thi-ciacs-99,cjn-fswrsp-99} asks for 
a short tour along which one mobile guard can see the entire region.
If the region is unknown in advance, we are faced with the
{\em online watchman problem}. For a simple polygon,
Hoffmann et al.\cite{hikk-pep-01}
achieve a constant competitive ratio of 26.5, while
Albers et al.~\cite{aks-eueo-99} show that no constant competitive
factor exists for a region with holes, and unbounded aspect ratio. 
Kalyanasundaram and Pruhs~\cite{kp-cctli-94} consider the problem in graphs
and give a competitive factor of 18.

In the context of geometric searching, a crucial issue is the 
question of how to look
around a corner: Given a starting position, and a known distance to
a corner, how should one move in order to see a hidden object
(or the other part of the wall) as quickly as possible?
This problem was solved by Icking et 
al.\cite{ikm-hlac-93,ikm-ocsla-94} who show that an optimal strategy
can be characterized by a differential equation that yields a competitive
factor of 1.2121\ldots, which is optimal. It should be noted
that actually using this solution requires 
numerical evaluation.

{\bf An Autonomous Mobile Robot.}
 From the practical side, our work is motivated by an actual application
in robotics:
The Fraunhofer Institute for Autonomous Intelligent Systems (AIS)
has developed autonomous mobile robots that can survey their environment
by virtue of a high-resolution, 3D laser scanner
\cite{RAAS2003}. By merging several 3D scans acquired in a stop, scan, plan, 
go fashion, the robot Kurt3D
builds a virtual 3D environment that allows it to navigate, avoid
obstacles, and detect objects\cite{IAS2004}. This makes the visibility problems
described above quite practical, as actually using good trajectories
is now possible and desirable.

However, while human mobile guards are
generally assumed to have full vision at all times,
our autonomous robot has to stop and take some time
for taking a survey of its environment. This makes the objective function
(minimize total time to locate an object or explore a region)
a sum of travel time and scan time; a somewhat related problem is searching for
an object on a line in the presence of turn cost~\cite{dfg-ostc-04},
which turns out to be a generalization of the classical linear search 
problem.
Somewhat surprisingly, scan cost (however small it may be) 
causes a crucial difference to
the well-studied case without scan cost, even in the limit of infinitesimally
small scan times.

{\bf Other Related Work.}
Visbility-based navigation of robots involves a variety of different
aspects. For example, Efrat et al.~\cite{egkp-ostcpt-03} study
the task of developing strategies for tracking and capturing
a visible target with known trajectory, while maintaining
line-of-sight among obstacles. Kutulakos et al.~\cite{kdl-psvgetd-94}
consider the task of vision-guided exploration,
where the robot is assumed to move about freely in three dimensions,
among various obstacles. 

{\bf Our Results.}
The main objective of this paper is to demonstrate that 
technology has reached the stage of actually
applying previous theoretical studies, at the same time
triggering new algorithmic research. We hope that this
will highlight the need for and the opportunities of closer 
interaction between theoreticians and practitioners. In particular,
we describe the problem of online searching by a real autonomous robot,
for an object (a chair) hidden behind a corner, which is at distance
$d$ from the robot's starting position. Our mathematical results
are as follows:

\vspace*{-.3cm}
\begin{itemize}
\item We show that for an initial distance of at least 
$d\geq 1/\delta$ from the corner, a competitive ratio of $2-\delta$ cannot
be achieved. This implies a lower bound of 2 on the competitive ratio
by any one strategy, and proves that there
is an important distinction from the case without scan cost.
\item We describe a heuristic strategy that is fast to evaluate and
easy to implement in real life.
\item We show that this strategy is asymptotically optimal
by proving that for large distances, the competitive ratio
converges to 2, matching the lower bound.
\item We give additional numerical evidence showing that the 
performance of our strategy is within about 2\% of the optimum.
\item Most importantly, we describe how our strategy can actually
be used by Kurt3D, a real mobile autonomous robot.
\end{itemize}

Further documentation of our work is provided by a video~\cite{fkn-sarv-04}
that is available at the authors' web address.

The rest of this paper is organized as follows. In Section~2, we describe
the technical details, properties, and capabilities of Kurt3D, an
autonomous mobile robot that was used in our experiments. 
Section~3 provides mathematical results
on the problem arising from Kurt searching for a hidden object.
Section~4 gives a description of how our results are used in practice.
The final Section~5 provides some directions for future research.

%% file: s2.tex
\section{The Autonomous Mobile Robot}

\vspace*{-3mm}
\subsection{The Kurt3D Robot Platform}

Kurt3D (Figure \ref{dddscanner}, top left) is a mobile robot
platform with a size of 45~cm (length) $\times$ 33~cm (width)
$\times$ 26~cm (height) and a weight of 15.6~kg. Equipped with
the 3D laser range finder the height increases to 47~cm and the
weight to 22.6 kg.\footnote{Videos of the exploration with the
autonomous mobile robot can be found at
\texttt{http://www.ais.fhg.de/ARC/kurt3D/index.html}} Kurt3D's
maximum velocity is 5.2~m/s (autonomously controlled
4.0~m/s). Two 90 W motors are used to power the 6 wheels, where
the front and rear wheels have no tread pattern to enhance
rotating. Kurt3D operates for about 4~hours with one battery (28
NiMH cells, capacity: 4500~mAh) charge. The core of the robot is
a Pentium-III-600~MHz with 384~MB RAM. An embedded 16-Bit CMOS
microcontroller is used to control the motor.

\vspace*{-3mm}
\subsection{The AIS 3D Laser Range Finder}\label{3dscanner}

\begin{figure}
\begin{center}
  \includegraphics[width=57.5mm,height=57.5mm]{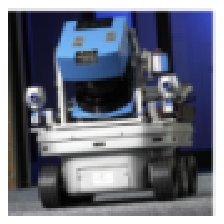}
  \includegraphics[width=57.5mm,height=57.5mm]{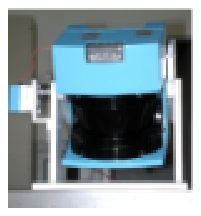}
  
\vspace*{1mm}

  \includegraphics[width=38mm,height=34mm]{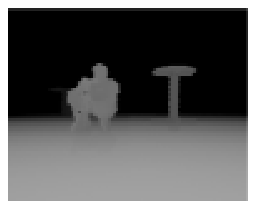}
  \includegraphics[width=38mm,height=34mm]{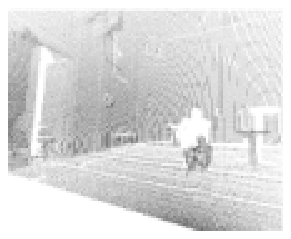}
  \includegraphics[width=38mm,height=34mm]{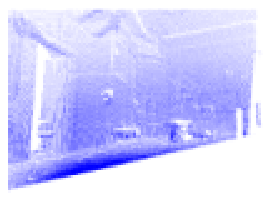}
\caption{Top left: The autonomous mobile robot Kurt3D equipped
with the 3D scanner. Top right: The AIS 3D laser range
finder. Its technical basis is a SICK 2D laser range finder
(LMS-200). Bottom row, left: A scanned scene as depth
image. Middle and right: Scanned scenes as point cloud viewed
with a camera orientation towards the door.}
\label{dddscanner}
\end{center}
\vspace*{-6mm}
\end{figure}

The AIS 3D laser range finder (Figure \ref{dddscanner}, top
right) \cite{RAAS2003, ISR2001} is built on the basis of a 2D
range finder by extension with a mount and a standard servo
motor. The 2D laser range finder is attached in the center of
rotation to the mount for achieving a controlled pitch
motion. The servo is connected on the left side (Figure
\ref{dddscanner}, top middle). The 3D laser scanner operates up
to 5h (Scanner: 17~W, 20 NiMH cells with a capacity of 4500~mAh,
Servo: 0.85~W, 4.5~V with batteries of 4500~mAh) on one battery
pack.

The area of $180^{\circ}\mbox{(h)} \times 120^{\circ}\mbox{(v)}$
is scanned with different horizontal (181, 361, 721 pts.) and
vertical (210, 420 pts.) resolutions. A plane with 181 data
points is scanned in 13~ms by the 2D laser range finder (rotating
mirror device). Planes with more data points, e.g., 361, 721,
duplicate or quadruplicate this time. Thus, a scan with 181
$\times$ 210 data points needs 2.8 seconds. In addition to the
distance measurement, the 3D laser range finder is capable of
quantifying the amount of light returning to the scanner, i.e.,
reflectance data \cite{IAS2004}. Figure \ref{dddscanner} (bottom
left) shows a scanned scene as depth image, created by off-screen
rendering from the 3D data points (Figure \ref{dddscanner}, bottom
middle) by an \texttt{OpenGL}-based drawing module.

\vspace*{-3mm}
\subsection{Basic 3D Scanner Software}

The basis of the scan matching algorithms and the reliable robot
control are algorithms for reducing points, line detection,
surface extraction and object segmentation. Next we give a brief
description of these algorithms. Details can be found in
\cite{ISR2001}.

The scanner emits the laser beams in a spherical way, such that
the data points close to the source are more dense. The first
step is to reduce the data. Therefore, data points located close
together are joined into one point. The number of these so-called
\emph{reduced points} is one order of magnitude smaller than the
original one.

\begin{figure}
\begin{center}
  \includegraphics[width=38mm,height=34mm]{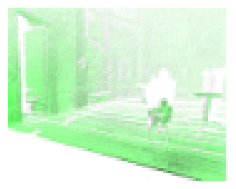}
  \includegraphics[width=38mm,height=34mm]{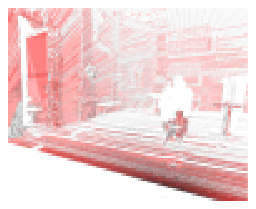}
  \includegraphics[width=38mm,height=34mm]{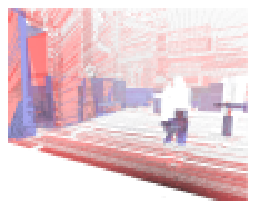}
\caption{Left: Line detection in every scan slice. Middle:
  Surface segmentation. Right: Bounding boxes of objects
  superimposing the surfaces}
\label{scannerbasic}
\end{center}
\end{figure}

Second, a simple length comparison is used as a line detection
algorithm. Given that the counterclockwise ordered data of the laser
range finder (points $a_0, a_1, \ldots, a_n$) are located on a
line, the algorithm has to check for $a_{j+1}$ if $ \| a_i,
a_{j+1} \| \ / \ \sum^j_{t=i}\| a_t,a_{t+1} \| < \epsilon(j)$ in order
to determine if $a_{j+1}$ is on line with
$a_j$. (Figure \ref{scannerbasic}, left)

The third step is surface detection. Scanning a plane surface,
line detection returns a sequence of lines in successive scanned
2D planes approximating the shape of surfaces. Thus a plain
surface consists of a set of lines. Surfaces are detected by
merging similar oriented and nearby
lines. (Figure \ref{scannerbasic}, middle)

The fourth and final step computes occupied space. For this
purpose, conglomerations of surfaces and polygons are merged
sequentially into objects. Two steps are necessary to find
bounding boxes around objects. First a bounding box is placed
around each large surface. In the second step objects close to
each other are merged together, e.g., one should merge objects
closer than the size of the robot, since the robot cannot pass
between such objects (Figure \ref{scannerbasic}, right). These
bounding boxes are used for avoiding obstacles.

Data reduction, line, surface and object detection are
real-time capable and run in parallel to the 3D scanning process.

\vspace*{-12mm}
\subsection{3D Scan Matching}

\begin{figure}
\begin{center}
  \includegraphics[width=38mm,height=34mm]{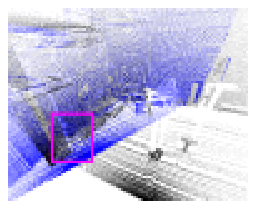}
  \includegraphics[width=38mm,height=34mm]{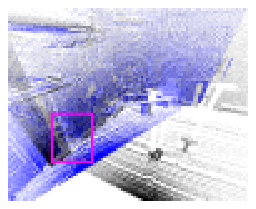}
  \includegraphics[width=38mm,height=34mm]{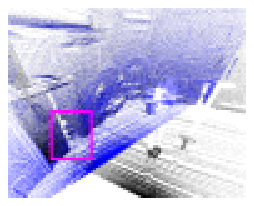}
\caption{Three iteration steps of scan alignment process for the
  two 3D scans presented in Figure \ref{dddscanner} (bottom, middle,
  right).}
\label{scanmatch}
\end{center}
\end{figure}

To create a correct and consistent representation of the environment,
the acquired 3D scans have to be merged in one coordinate
system. This process is called registration. Due to the robot's
sensors, the self-localization is usually erroneous and
imprecise, so the geometric structure of overlapping 3D scans has
to be considered for registration. The odometry-based robot pose
serves as a first estimate and is corrected and updated by the
registration process. We use the well-known Iterative Closest
Points (ICP) algorithm \cite{Besl_1992} to compute the
transformation, consisting of a rotation $\M R \in \RR^{3 \times
3}$ and a translation $\V t \in \RR^3$.  The ICP algorithm
computes this transformation in an iterative fashion. In each iteration
the algorithm selects the closest points as correspondences
and computes the transformation ($\M R, \V t$) for minimizing
\begin{eqnarray*}
E(\M R, \V t) \! = \! \sum_{i=1}^{N_m}\sum_{j=1}^{N_d}w_{i,j}\norm{\V
  m_{i}-(\M R
\V d_j+\V t)}^2, \label{DMin}
\end{eqnarray*}
where $N_m$ and $N_d$ are the number of points in the model set
$M$, i.e., first 3D scan, or data set $D$, second 3D scan,
respectively, and $w_{ji}$ are the weights for a point match. The
weights are assigned as follows: $w_{ji} = 1$, if $\V m_i$ is the
closest point to $\V d_j$ within a close limit, $w_{ji} = 0$
otherwise. It is shown in \cite{Besl_1992} that the iteration 
terminates in a minimum. 
The assumption is that in the last iteration
the point correspondences are correct.  In each iteration
the transformation is computed in a fast closed-form manner by
the quaternion-based method of Horn \cite{Horn_1987}. In addition,
point reduction and $k$D.trees speed up the computation of the
point pairs, such that only the time required for scan matching is
reduced to roughly one second
\cite{RAAS2003}. Figure \ref{scanmatch} shows three iteration steps
for 3D scan alignment.

\vspace*{-12mm}
\subsection{3D Object Detection}

\begin{figwindow}[0,r,{
\includegraphics[width=38mm,height=32mm]{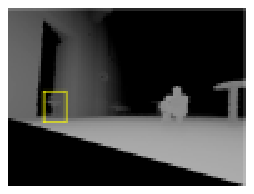}},{Object
	 detection in range images.\label{chair}}]
\noindent
Automatic, fast and reliable object detection algorithms are
essential for mobile robots used in searching tasks. To perceive
objects, we use the 3D laser range and reflectance data. The 3D
data is transformed into images by off-screen rendering. To
detect objects, a cascade of classifiers, i.e., a linear decision
tree, is used.  Following the ideas of Viola and Jones, we
compose each classifier from several simple classifiers, which in
turn contain an edge, line or center surround feature
\cite{Viola_2001}. There exists an effective method for the fast
computation of these features using an intermediate
representation, namely, integral image. For learning of the
object classes, a boosting technique, namely, Ada Boost, is used
\cite{Viola_2001}. The resulting approach for object
classification is reliable and real-time capable and combines
recent results in computer vision with the emerging technology of
3D laser scanners. For a detailed discussion of object detection
in 3D laser range data, refer to
\cite{IAS2004}. Figure \ref{chair} shows an office chair detected
by a cascade of classifiers.
\end{figwindow}

\bigskip

%% file: s3.tex
\section{Algorithmic Approach}
When trying to develop a good search strategy,
we have to balance theoretical quality with
practical applicability. More precisely, we have to
keep a close eye on the trade-off between these objectives:
An increase in theoretical quality may come at the
expense of higher mathematical difficulty, possibly requiring
more complicated tools.
In an online context, the use of such tools may cause both
theoretical and practical difficulties:
Complicated solutions may cause computational overhead that
can change the solution itself by causing extra delay;
on the practical side, actually applying such a solution
may be difficult (due to limited accuracy of the robot's motion)
and without significant use.
To put relevant error bounds into perspective:
The largest room available to us is the great hall of Schloss
Birlinghoven; even there, the size of Kurt and the object 
is still in the order of 2\% of the room diameter.

On the mathematical side, it should be noted that even in
the theoretical paper \cite{hikk-pep-01},
semi-circles are considered instead of the solution to the
differential equation, in order to allow analysis of the resulting
trajectories.

In the following, we will start by giving some basic mathematical
observations and properties (Section~\ref{subsec:basic});
this is followed by a discussion
of globally optimal strategies (Section~\ref{subsec:global}).
Section ~\ref{subsec:circle} describes a natural heuristic solution
that is both easy to describe and fast to evaluate; we give a number
of computational and empirical results that suggest our heuristic
is within 2\% of an optimal strategy.
Finally, Section ~\ref{subsec:asymp} provides a number of
mathematical results, showing that our fast and easy
heuristic is asymptotically optimal.

\subsection{Basic Observations}
\label{subsec:basic}

First, we introduce some notation that will be used throughout this section.

Let us assume that the corner that hides the object
is at distance $d$ from the start.
Let $x_i$ denote the distance the robot travels in the
$i-$th step, i.e., on its way from position $i-1$ to position $i$, 
from which the $i-$th scan will be taken.
If the object was hidden infinitesimally behind position~$i$,
the optimal solution would go perpendicularly to the line $L_i$ that runs
from the corner through position $i$, and then take one scan from there.
Let $d_i$ denote the length of this line segment (observe that it meets $L_i$
at a point that lies on the semi-circle spanned by the start
and the corner).
Then the optimum cost to detect the object would be $1+d_i$, whereas the robot
would only see the object at position $i+1$, having accumulated a cost of
\[
    i+1 + \sum_{j=1}^{i+1} x_j.
\]
Now suppose that $c$ is the smallest competitive ratio that can be achieved
in this setting. By local optimality, for any scan position, the ratio of the
solution achieved and the optimal solution must be equal to $c$.
Therefore,
\begin{equation}
    x_{i+1} = c(1+d_i) - (i+1) - \sum_{j=1}^i x_j
\label{eq:rec}
\end{equation}
must hold for $i=1,2,\ldots$ In particular, we have $x_1 = c-1$ for the
first step.

\subsection{Globally Optimal Strategies}
\label{subsec:global}

The above recursion can be used for proving a lower bound.
\begin{theorem}
\label{th:lower}
There is no global $c$-competitive strategy with $c<2$.
\end{theorem}

\proof
Assume the claim was false, and there was a $c$-competitive
strategy for $c=2-\delta$. 
We show that $x_i \leq (1-\delta)^i$ holds,
making it impossible for the robot to get further than a distance
of $1/\delta$ away from the start, a contradiction.
Clearly, we have $x_1=1-\delta$ for step~1.
Moreover,
\[
     d_i \leq \sum_{j=1}^i x_j
\]
holds, because $d_i$ is the shortest path from the start to line $L_i$,
whereas the sum denotes the length of the robot's path.
Plugging this into our recursion yields
\[
    x_{i+1} \leq (1 - \delta) (1+\sum_{j=1}^i x_j) -i.
\]
By induction, we have $x_j \leq (1-\delta)^j$, hence
\begin{equation*}
   x_{i+1} \leq  (1 - \delta)\frac{1-(1-\delta)^{i+1}}{\delta} -i \leq 
   (1-\delta)^{i+1},
\end{equation*}
using the Bernoulli inequality $1-(1+i)\delta \leq (1-\delta)^{i+1}$.
\qed

\medskip
Instead of increasing the distance $d$ we could as well
consider a situation where start and corner are a distance~1 apart, but the
scan cost is only $1/d$. Now Theorem~\ref{th:lower}
shows a remarkable discontinuity: Even for a scan cost arbitrarily small,
a lower bound of~2 cannot be beaten, whereas for zero scan cost, a factor
of $1.212\dots$ can be obtained~\cite{ikm-hlac-93}.

On the positive side, for $n$ intermediate scan points, Equation~(\ref{eq:rec})
provides $n$ optimality conditions. As there are $2n$ degrees
of freedom (the coordinates of intermediate scan points), 
we get an underdetermined nonlinear optimization problem for any given 
distance $d$, provided that we know the number of scan points. 
For $d=1$, this can be used to derive an optimal competitive factor of
1.808201..., achieved with one intermediate scan point.
For larger $d$ (and hence, larger $n$) one could derive additional
geometric optimality conditions and use them in combination with more 
complex numerical methods.
However, this approach appears impractical for real applications,
for reasons stated above. As we will see in the following, there is
a better approach.

\subsection{A Simple Heuristic Strategy}
\label{subsec:circle}

Now we describe a simple strategy for the searching
problem that uses trajectories inscribed into a circle. This
reduces the degrees of freedom to the point where evaluation
is fast and easy. What is more, it works very
well in realistic settings, and it is asymptotically optimal for
decreasing cost of scanning, or growing size of the environment.

\begin{figure}[htbp]
\centerline{\epsfig{file=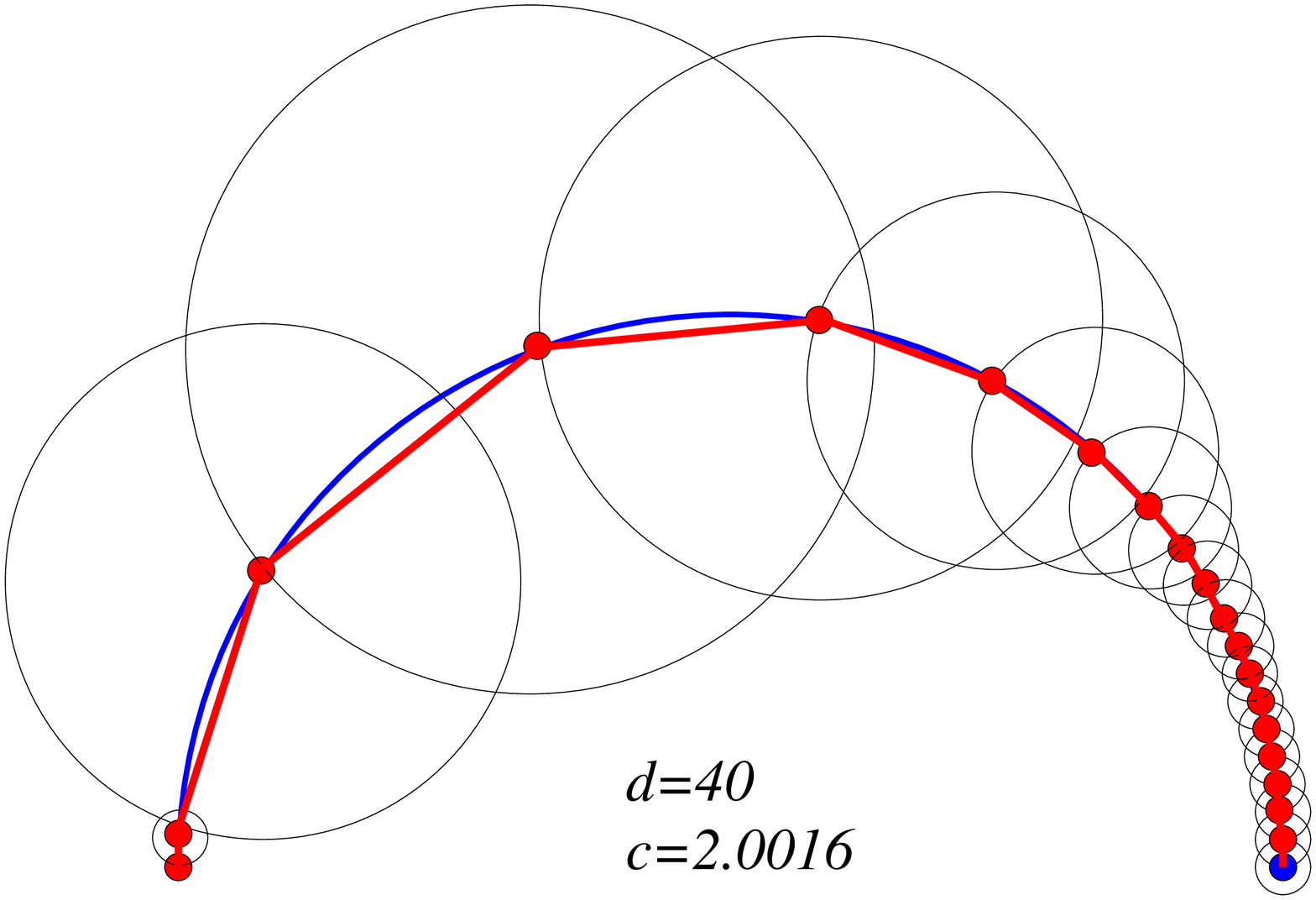,width=0.6\textwidth}}   
\centerline{\epsfig{file=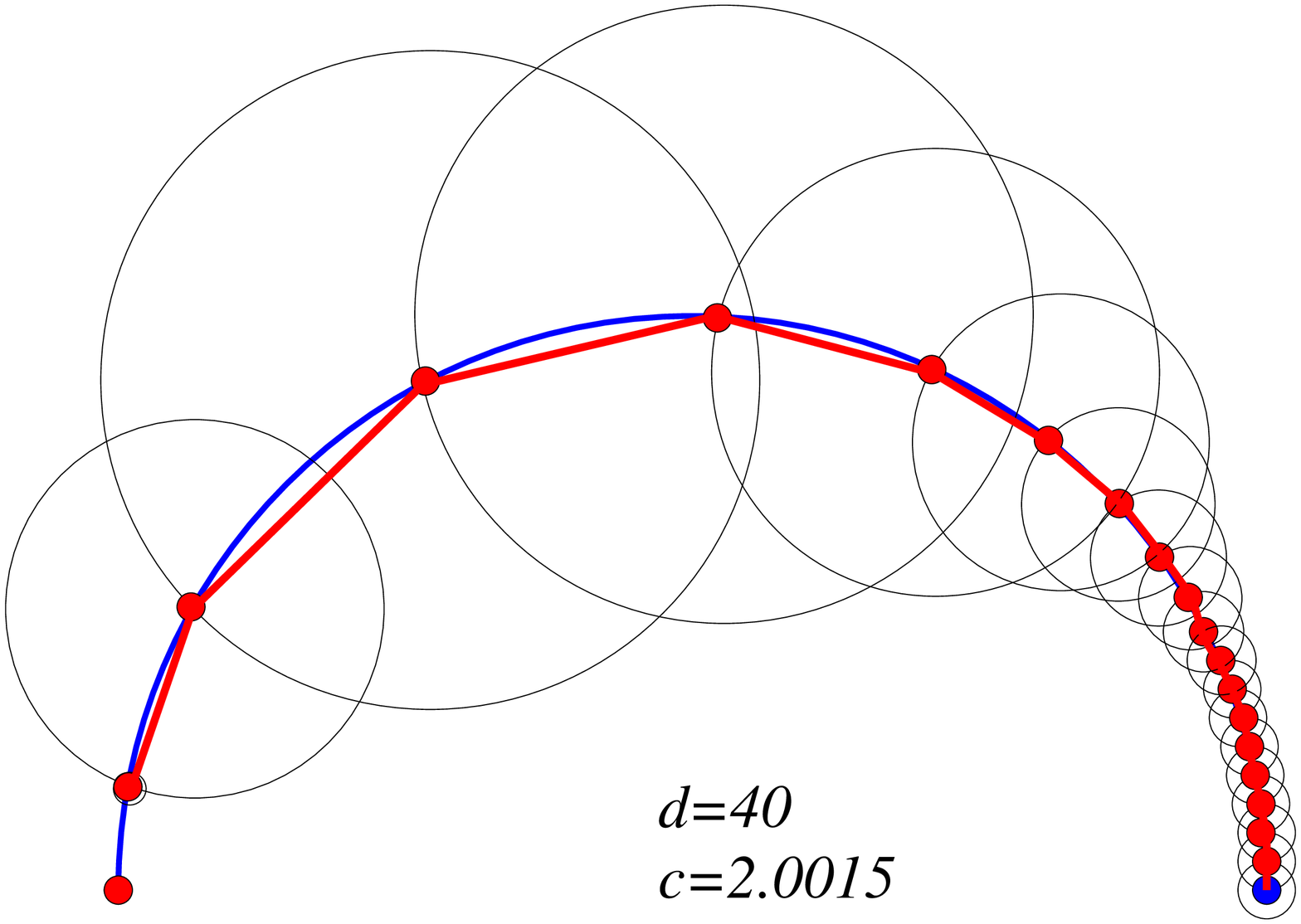,width=0.6\textwidth}}   
\label{fig:40}
\caption{An example for $d=40$: (top) For $c=2.0016$, the circle sequence
reaches the corner, showing that the chosen $c$ can be achieved.
(bottom) For $c=2.0015$, the sequence collapses before reaching
the corner, showing that the chosen $c$ cannot be achieved.
The actual optimum is about 2.001525...}
\end{figure}

The robot simply follows a polygonal path inscribed into
the semi-circle of diameter $d$, spanned by start
and corner. It remains to determine those points where it
stops for scanning its environment. This is done by applying the
optimality condition 
derived in Section~\ref{subsec:basic}.
In step $j$, the robot moves along a chord of 
length $x_j$. From the corner, this chord is visible
under an angle of $\varphi_j= \arcsin(x_j/d)$.
The chord connecting the start to position $i$ is of length
\[
    d_i = d\ \sin( \sum_{j=1}^i \varphi_j),
\]
so that the recursion (\ref{eq:rec}) 
obtained in Section~\ref{subsec:basic} turns into
\begin{eqnarray*}
x_{i+1}&=&c\left(1+d\ \sin\left(\sum_{j=1}^{i}\arcsin\left(\frac{x_j}{d}\right)\right)\right)
    -(i+1)-\sum_{j=1}^{i}x_j.
\end{eqnarray*}
Given any $c>1$, we can tentatively compute steps of length $x_i$
by this formula, starting with $x_1=c-1$.
If the resulting sequence reaches the corner, 
the ratio of $c$ can indeed be achieved.
If it collapses prematurely (by returning negative values) $c$ was too small.
(For example, $c=2.001525...$ is optimal for $d=40$; see Figure~\ref{fig:40}
for an illustration of upper and lower bounds on this value.)

\begin{figure}[htbp]
\centerline{\epsfig{file=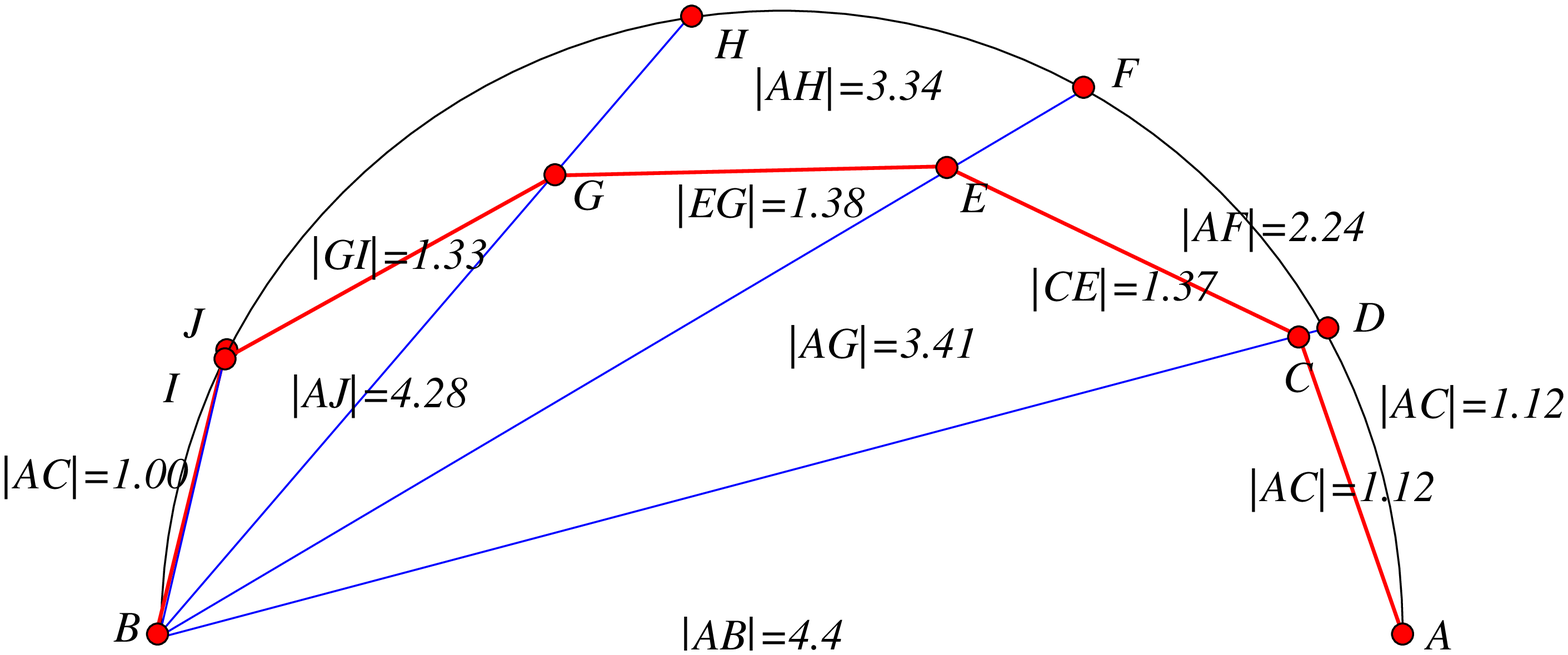,width=0.8\textwidth}}   
\caption{A solution for $d=4.4$ that achieves competitive ratio 2.12:
The starting position is at $A$, the corner at $B$.}
\label{fig:212}
\end{figure}

\begin{figure}[htbp]
\centerline{\epsfig{file=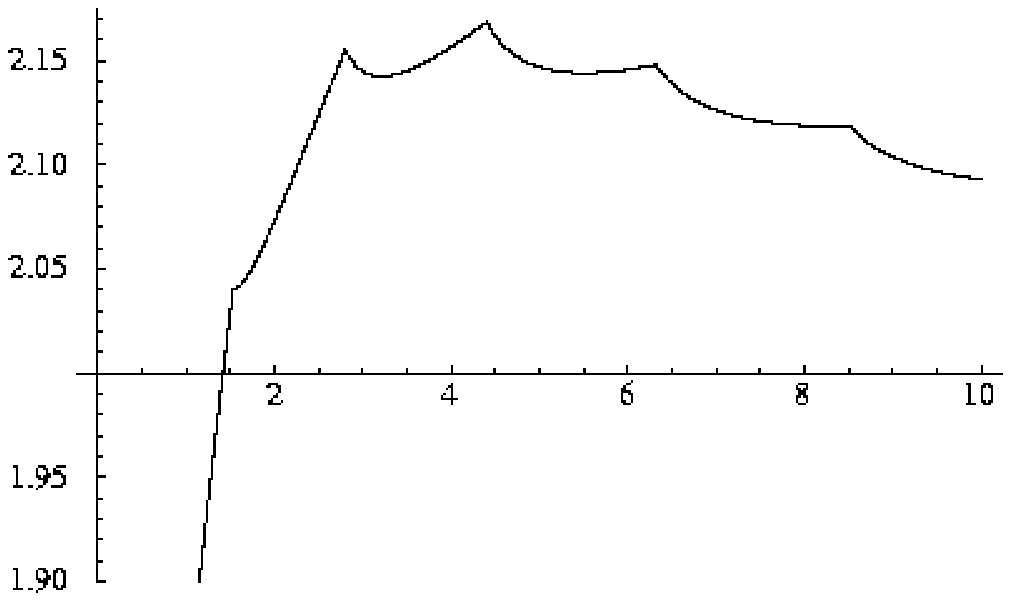,width=0.8\textwidth}}   
\vspace*{5mm}
\centerline{\epsfig{file=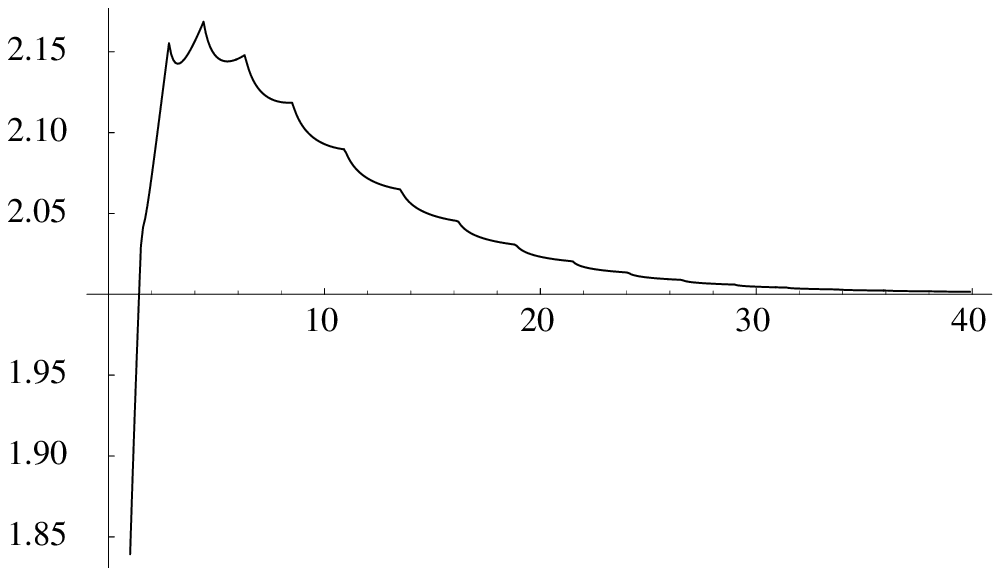,width=0.8\textwidth}}   
\caption{The competitive ratio as a function of $d$:
(top) for small values of $d$;
(bottom) for larger values of $d$;
note the cusps at threshold values, the sharp peak at about (4.4,2.17), and
the clear asymptotic behavior. The first step length, $x_1$ is given by
$c-1$.}
\label{fig:func}
\end{figure}

By performing a binary search, the optimal ratio and the necessary
step lengths can be computed extremely fast. Moreover, an analysis of
the optimal ratio as a function of $d$ shows that a maximum is reached
for $d=4.400875...$
which is precisely at the threshold between three and four necessary scans,
with a competitive ratio of 2.168544.
(See Table~\ref{tab:thresh} for an overview of the critical values
for which the number of scans increases, and Figure~\ref{fig:func}
for the achievable ratios as a function of the distance.)
This is still within about 2\% of the global optimum, which appears to be
at about 2.12 (see Figure~\ref{fig:212}.)
Moreover, numerical evidence shows
that the ratio approaches 2 quite rapidly
as $d$ tends to infinity. This is all the more surprising, as the resulting
initial step length converges to 1, while a constant step length
of 1 yields a competitive ratio of $\pi$.
In the following Section~\ref{subsec:asymp} we give a mathematical
proof of this observation.

\begin{table}[t!]
\begin{center}
\noindent
\begin{tabular}{|c|c|c|}
\hline
\ \ Number    \ \   & \ \ Maximal  \ \ & \ $c$ at \ \\
\ \ of scans  \ \   & \ \  $d$     \ \ & \ upper bound \ \\
\hline
\hline
0 & 0.618034& 1.618034\\
1 & 1.530414& 2.040287\\
2 & 2.799395& 2.155363\\
3 & 4.400876& 2.168544\\
4 & 6.316892& 2.147994\\
5 & 8.514200& 2.118498\\
\hline
\end{tabular}
\caption{Threshold values for small numbers of scans, rounded to six digits.}
\label{tab:thresh}
\end{center}
\end{table}

\subsection{Asymptotics}
\label{subsec:asymp}

As we have seen in Theorem~\ref{th:lower}, there is a
lower bound of 2 on the competitive ratio for all strategies and large $d$.
In the following we will show that for large $d$, there is a matching
upper bound on our circle strategy presented in Section~\ref{subsec:circle},
proving it to be asymptotically optimal. For limited physical distances, 
it shows that even for arbitrarily small scan times, there is a
relatively simple strategy that achieves the optimal ratio of 2.

Our proof of the upper bound proceeds as follows. Let us assume that we are given
some fixed $\varepsilon>0$.
We then proceed to show that
for $c=2+\varepsilon$, the recursion presented in Section~\ref{subsec:circle}
does not collapse before the corner is reached, if the diameter $d$ of the 
semi-circle is large enough.

In proving the lower bound
stated in Theorem~\ref{th:lower}, we have used the obvious fact that 
the length $d_n$ of the optimal path cannot exceed the length of the robot's 
path. Now we are turning this argument around: The robot's path to position $n$
does not exceed the length of the circular arc leading from the start to
position $n$. As this arc is not much longer than 
$d_n$, the length of the chord
from the start to $n$, if the diameter $d$ of the circle is large enough.
More precisely, we use the following.

\begin{lemma}
\label{le:arcchord}
(i) There is an upper bound on the total length of the first $n$ steps
of the circle strategy that does only depend on $n$ and $\varepsilon$,
but not on $d$.\\
(ii) Given any $A>0$, we can find  $d_0$ such that each 
arc of length $\leq A$ in a circle of diameter $\geq d_0$ exceeds the 
length of its chord by at most $\varepsilon^2$.
\end{lemma}

\proof
Claim (i) can be shown by the same technique as in the proof of
Theorem~\ref{th:lower}. In order to prove claim~(ii), let $a$ and $c$
denote the maximum lengths of an arc and its chord in a circle of
diameter $d$ satisfying $a \leq b + \varepsilon^2$. Let $2\beta$ denote
the angle of the arc, as seen
from the center, so that $a=d \beta$ and $c=d \sin \beta$ hold. The maximum arc
satisfying the condition is of length $a=d \beta_d$ where $\beta_d$ is the
solution of the equation $\beta_d - \sin \beta_d = \varepsilon^2 /d$. In the
equivalent expression
\[
    \ d \beta_d\ \left(1 - \frac{\sin \beta_d}{\beta_d} \right) = \varepsilon^2
\]
the fraction tends to one, so $a= d \beta_d$ must be unbounded.
\qed

These facts will now be used in providing a lower bound for the first steps
along the semi-circle, aiming for a competitive
ratio of $c=2+\varepsilon$.

\begin{lemma}
\label{le:liftoff}
Let $\varepsilon>0$ and $N$ be given. Then there is a
number $d_0$ such that for each diameter $d\geq d_0$ we have
$x_n\geq 1+\left(2^{n}-1\right)\varepsilon$, for
$n\leq N$.
\end{lemma}

\proof
Using Lemma~\ref{le:arcchord} we can choose $d_0$ large enough that

\[
   \sum_{i=1}^{n} x_i \leq d_n + \varepsilon^2
\]
holds for all $n \leq N$ if $d \geq d_0$.
Now we proceed by induction. For $x_1:=(1+\varepsilon)$
the claim is fulfilled. For $n=2$ we observe that
$d_1 = x_1$ holds, so the recursive formula~(\ref{eq:rec})
yields
\begin{eqnarray*}
   x_2 &=& (2+\varepsilon) (1+d_1) - 2 - x_1 \\
       &=& (2+\varepsilon)^2 -3 -\varepsilon \geq 1+3\varepsilon.
\end{eqnarray*}
Now assume the claim was true for $x_1,\ldots,x_{n-1}$, where $n\geq 3$,
and let $d_{n-1}$ be the $(n-1)$st chord, arising by connecting
the start point with the $(n-1)$st scan point.
The induction hypothesis implies
$$\sum_{i=1}^j x_i
\geq \sum_{i=1}^j \left( 1+\left(2^{i}-1\right)\varepsilon\right)
= j + (2^{j+1}-j-2)\varepsilon.$$
 From the recursion we obtain

$$x_n=(2+\varepsilon)(1+d_{n-1})-n-\left(\sum_{i=1}^{n-1} x_{i}\right).$$

By choice of $d$
we have $d_{n-1}\geq \left(\sum_{i=1}^{n-1} x_{i}\right)-\varepsilon^2$
for $n\leq N$.
Thus, we get
\begin{eqnarray*}
x_n&\geq& (2+\varepsilon)\left(1+\left(\sum_{i=1}^{n-1} x_{i}\right)-\varepsilon^2\right)
           - n- \left(\sum_{i=1}^{n-1} x_{i}\right)\\
   &=& (1+\varepsilon)\left(\sum_{i=1}^{n-1} x_i \right) +
           (2+\varepsilon)(1-\varepsilon^2)-n\\
   &=& 1+(2^n-1)\varepsilon + (2^n -n-3-\varepsilon)\ \varepsilon^2 \\
   &\geq& 1+(2^n-1)\varepsilon,
\end{eqnarray*}
as $n\geq 3$.
\qed

\medskip
Under the assumptions of Lemma~\ref{le:liftoff} we
can now prove the following.
\begin{lemma}
\label{le:rise}
For the first $N$ steps of the robot,
$\frac{1}{N}\sum_{i=0}^{N-1} x_i\geq 5$ holds.
\end{lemma}

\proof
We may assume that
$$x_n\geq 1+\left(2^{n}-1\right)\varepsilon$$
holds for $n \leq N$. If $N$ is large enough and $n\geq N/2$, we get
$$ x_{n}\geq  1+ \left(\frac{10}{\varepsilon}-1\right)\varepsilon \geq 10.$$
Thus,
$$\sum_{i=1}^{N-1} x_i\geq \sum_{i=N/2}^{N-1} x_i\geq 5N,$$
as claimed.
\qed

\medskip
To conclude the proof, we consider a diameter $d$ large
enough for Lemma~\ref{le:rise} to hold, so we have a lower
bound of 5 on the average size for the first $N$ steps.
This suffices to show that all following steps are at
least of length 5.

\begin{lemma}
\label{le:glide}

Assume that for some $N\geq 12$, we have
$\sum_{i=1}^{N-1} x_i\geq 5N$.
Then $x_n\geq 5$ for all $n\geq N$.
\end{lemma}

\proof
Again we proceed by induction and consider

$$x_n=(2+\varepsilon)(1+d_{n-1})-n-\left(\sum_{i=1}^{n-1} x_{i}\right).$$
As all $x_i$ are lengths of chords of the semi-circle
with diameter $d$, we have

$$d_{n-1}\geq \frac{2}{\pi}\sum_{i=1}^{n-1} x_i.$$
By a similar argument as before, we get

\begin{eqnarray*}
x_n&\geq& (2+\varepsilon)\left(1+\frac{2}{\pi} \left(\sum_{i=1}^{n-1} x_{i}\right)\right)- n
            - \left(\sum_{i=1}^{n-1} x_{i}\right)\\
   &\geq& \left(\frac{4}{\pi}-1\right)\left(\sum_{i=1}^{n-1} x_{i}\right) - n+2\\
   &\geq& \left(\frac{4}{\pi}-1\right)\ 5n - n+2 \geq 5,
\end{eqnarray*}
since $n\geq 12$, as claimed.
\qed

\medskip
With the help of these lemmas, we get

\begin{theorem}
\label{th:2asymp}
The circle strategy is asymptotically optimal:
For any $\varepsilon>0$, there is a $d_\varepsilon$, such that
for all $d\geq d_\varepsilon$, the strategy is $(2+\varepsilon)$-competitive.
\end{theorem}

\proof
The preceding Lemmas~\ref{le:liftoff}, \ref{le:rise}, \ref{le:glide}
show that for any large enough $d$, the sequence will consist of step lengths
that are all at least 5. This implies that the sequence will reach the
corner in a finite number of steps, showing that a competitive factor
of $(2+\varepsilon)$ can be reached.
\qed

%% file: s4.tex
\section{Practical Application}

Our strategy was used in a practical setting,
documented in the video~\cite{fkn-sarv-04}.
In the great hall of Schloss Birlinghoven, starting about 8 meters from a door
($d=1$ for the right scanner setting), Kurt follows the trajectory 
developed in the third part; depending on the position of a hidden object
(a chair) he may have to perform a second scan from the corner.
The second scenario shows a starting distance of $d=2$, resulting in
two intermediate scan points.

%% file: s5.tex
\section{Conclusions}

\begin{center}
\begin{figure}[htbp]
  \includegraphics[width=.49\textwidth]{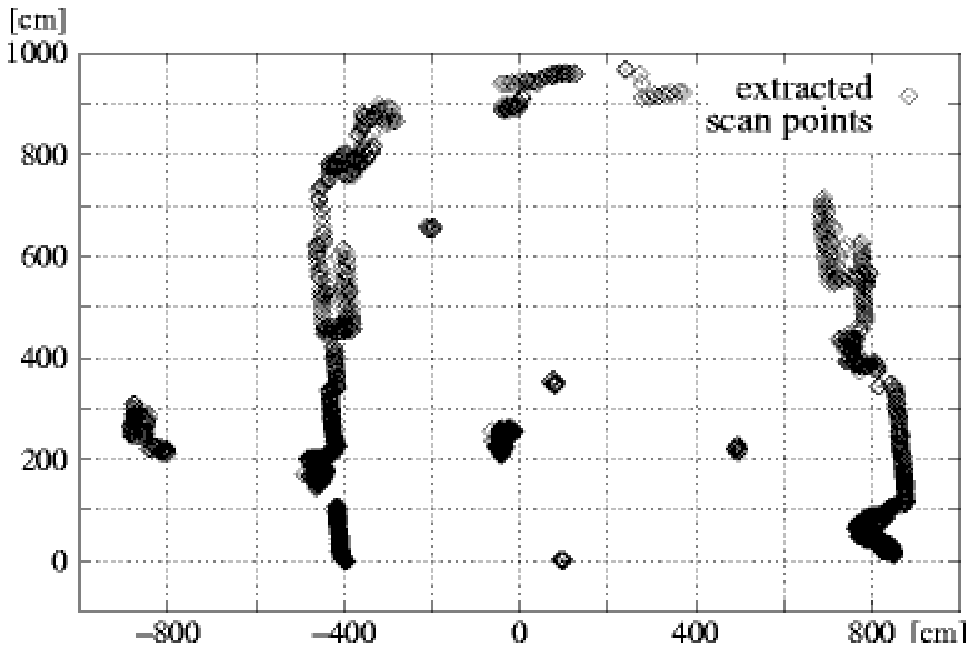}
  \includegraphics[width=.49\textwidth]{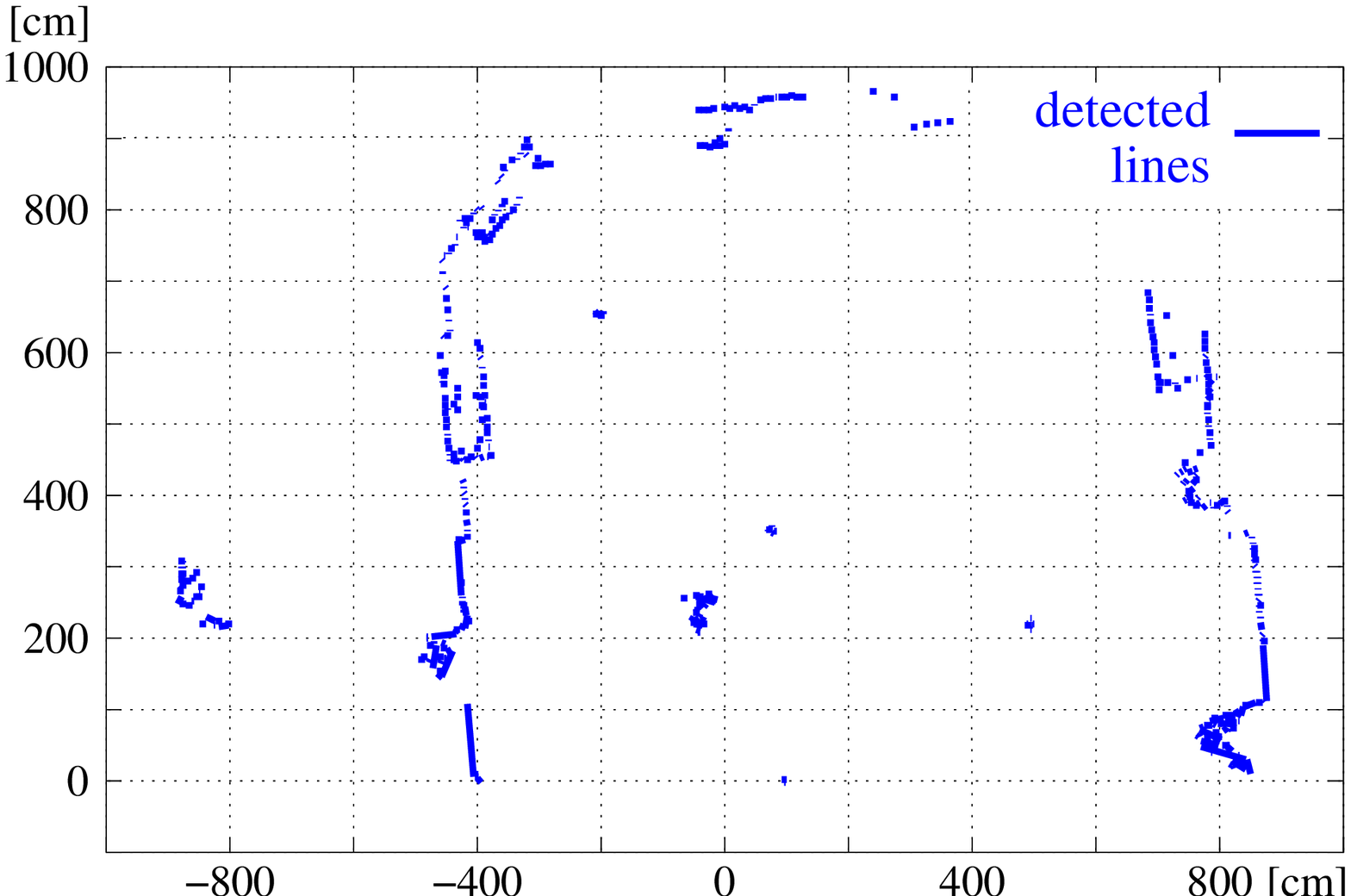}\\
  \hspace*{.25\textwidth}
  \includegraphics[width=.49\textwidth]{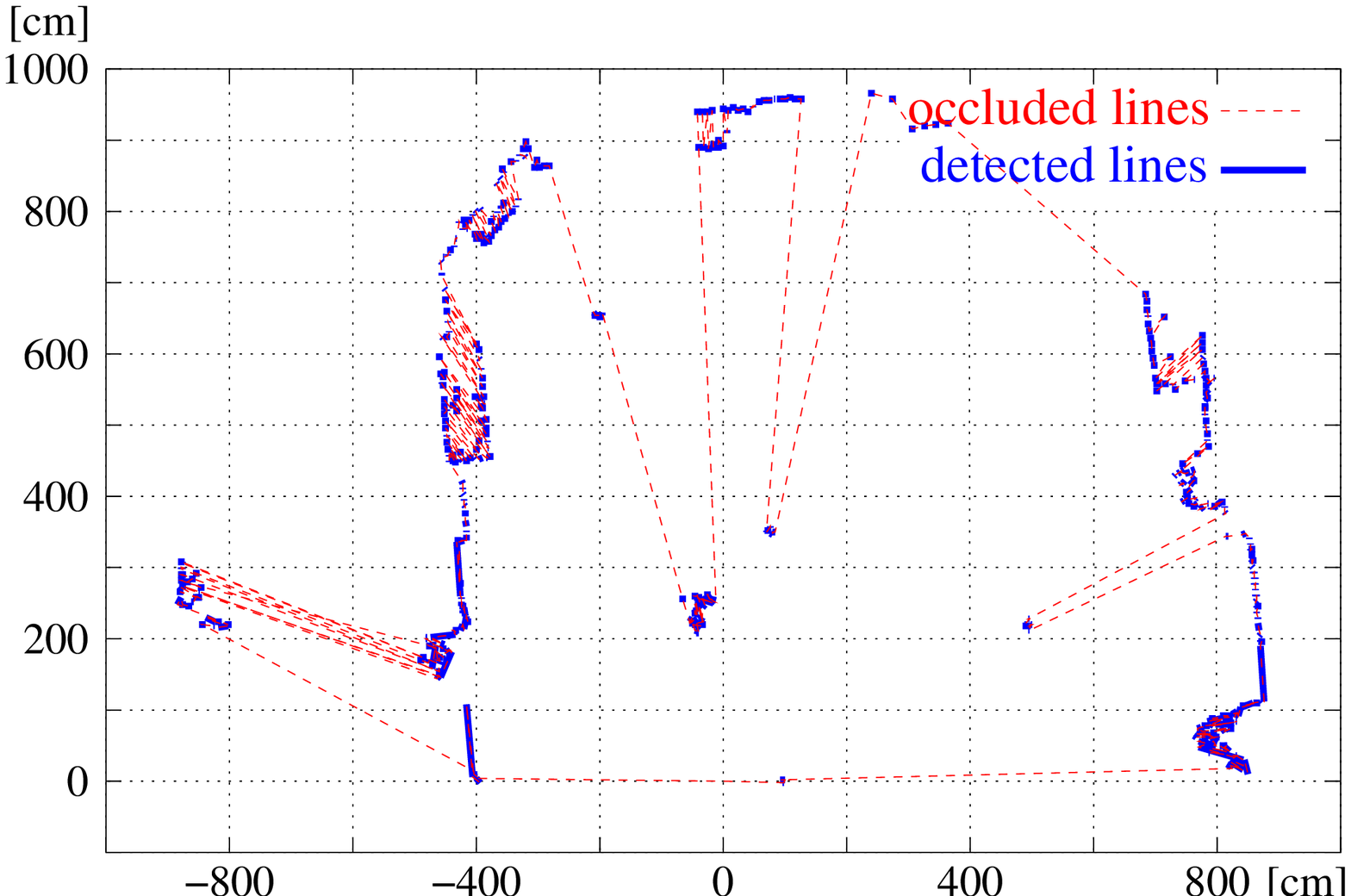}
\caption{A typical scenario faced by Kurt3D. Top left: Extracted
  points at height 75~cm (corresponding to figure
  \ref{dddscanner}, bottom middle). Top right: Line detection using
  Hough transform.  Bottom: Automatically generated map with
  occlusion lines \cite{RAAS2003}.}
\label{fig:riss}
\end{figure}
\end{center}

We have developed a search strategy that can be used for an actual
autonomous robot. Obviously, a number of problems remain.
Just like \cite{ikm-hlac-93} provided a crucial step
towards the solution for exploring general simple polygons
described in \cite{hikk-pep-01}, 
one of the most interesting challenges is to extend our results to
more general settings with a larger number of obstacles,
or the exploration of a complete region. See Figure~\ref{fig:riss}
for a typical realistic scenario.
It should be noted that scan cost (and hence positive step length
without vision) can cause theoretical problems in the presence of tiny
bottlenecks; even without scan cost, this is the basis of
the class of examples in \cite{aks-eueo-99} for polygons with holes. 
However, in a practical setting, lower bounds on the feature size are 
given by robot size and scanner resolution. Thus, there may be some hope.

\vspace*{-.5cm}

\vspace*{-1.4cm}
\section*{Acknowledgments}
This research was motivated by the Dagstuhl workshop on robot navigation,
Dec 7-12, 2003. We thank all other participants for a fruitful
atmosphere and motivating discussions. We thank 
Hartmut Surmann, Joachim Hertzberg, Kai Lingemann, Kai Perv\"olz, 
Matthias Hennig, Erik Demaine, Shmuel Gal, 
Christian Icking, Elmar Langetepe, Lihong Ma for preceding joint research
that laid the foundations for this work, and Matthias Hennig,
Rolf Mertig, and Jan van der Veen for technical assistance.